\documentclass[10pt,twocolumn,letterpaper]{article}

\usepackage[pagenumbers]{cvpr} % To force page numbers, e.g. for an arXiv version

\usepackage{graphicx}
\usepackage{amsmath}
\usepackage{amssymb}
\usepackage{booktabs}
\usepackage{adjustbox}

\usepackage[pagebackref,breaklinks,colorlinks]{hyperref}

\usepackage[capitalize]{cleveref}
\crefname{section}{Sec.}{Secs.}
\Crefname{section}{Section}{Sections}
\Crefname{table}{Table}{Tables}
\crefname{table}{Tab.}{Tabs.}

\def\cvprPaperID{5691} % *** Enter the CVPR Paper ID here
\def\confName{CVPR}
\def\confYear{2023}

\begin{document}

% \title{\LaTeX\ Author Guidelines for \confName~Proceedings}
\title{Two-in-one Knowledge Distillation for Efficient Facial Forgery Detection}
\author{
Chuyang Zhou$^1$,
Jiajun Huang$^1$, 
Daochang Liu$^1$, 
Chengbin Du$^{1}$, 
Siqi Ma$^2$,
Surya Nepal$^3$, 
Chang Xu$^1$ 
\\
$^1$School of Computing Science, University of Sydney\\
$^2$School of Engineering and Information Technology, UNSW Canberra, 
$^3$CSIRO Data61\\
\{czho4938@uni., jhua7177@uni., daochang.liu@, chdu5632@uni., c.xu@\}sydney.edu.au,\\
siqi.ma@adfa.edu.au, 
surya.nepal@data61.csiro.au}
\maketitle

%%%%%%%%% ABSTRACT
\begin{abstract}
Facial forgery detection is a crucial but extremely challenging topic, with the fast development of forgery techniques making the synthetic artifact highly indistinguishable.
Prior works show that by mining both spatial and frequency information the forgery detection performance of deep learning models can be vastly improved. However, leveraging multiple types of information usually requires more than one branch in the neural network, which makes the model heavy and cumbersome. 
Knowledge distillation, as an important technique for efficient modelling, could be a possible remedy.
We find that existing knowledge distillation methods have difficulties distilling a dual-branch model into a single-branch model. More specifically, knowledge distillation on both the spatial and frequency branches has degraded performance than distillation only on the spatial branch. 
To handle such problem, we propose a novel two-in-one knowledge distillation framework which can smoothly merge the information from a large dual-branch network into a small single-branch network, with the help of different dedicated feature projectors and the gradient homogenization technique.
Experimental analysis on two datasets, FaceForensics++ and Celeb-DF, shows that our proposed framework achieves superior performance for facial forgery detection with much fewer parameters.
\end{abstract}
%%%%%%%%% BODY TEXT

\section{Introduction}
\label{sec:intro}

% Forgery is a concern, forgery data, Intro GAN technology — hard to distinguish by human eyes
% Intro of forgery data generation
% How SOTA discrimination models work (multi-branch), reason that sota model use RGB and Frequency.
% Problem of SOTA models… so that we need model compression
% We tried existing KD models, but not good enough (problem in merging different types of information)
% So we propose the two-in-one framework, outperform all the existing methods.

% social concerns about abusing these techniques also rapidly increased. This brings out researches about facial forgery detection which is an extremely challenging topic since the latest technologies can easily cheat the human eyes. 
% for dual-branch models that leverage both RGB and frequency information,
% Although these methods achieved promising performance on discriminating the forgery and pristine data,

% \begin{figure}[t]
%   \centering
%   \includegraphics[width=0.9\linewidth]{latex/img/intro.pdf}

%   \caption{The proposed TOKD framework distills complementary information from both spatial and frequency domains for efficient facial forgery detection, with a gradient homogenization mechanism reducing the conflicts between the two domains.}
%   \label{fig:intro}
% \end{figure}

% Concern about forgery data
The development of deep neural networks dramatically benefits image forging. Techniques such as the deep generative models can not only manipulate the elements of images, but also generate brand new synthetic images from scratch. Among all kinds of image synthesis applications, facial forgery grabs plenty of social attention since it can pose significant risk to digital information security. Recent advances in facial image synthesis, stemming from Generative Adversarial Network (GAN)\cite{goodfellow2020generative} to recent developments including DeepFake\cite{GitHubDFfaceswap}, FaceSwap\cite{GitHubMKfaceswap}, Face2Face\cite{thies2016face2face} and NeuralTextures\cite{thies2019deferred}, can forge the human face medias with superior quality and naturalness. 
These methods can produce realistic medias in which almost all the forgery artifacts are concealed, which means it is nearly impossible to distinguish the pristine images and forgery images only through human visual perception.
This kind of forgery data generation techniques can be maliciously used for identity cheating, rumor spreading and can even lead to crisis of confidence. 
Therefore, the research of more effective forgery detection method is of great importance.

% Intro of forgery data generation

% How latest forgery detection models work, from naive to complex methods, multi-branch in the end 
% Add some reference
To deal with this problem, many facial forgery detection methods relying on the deep neural networks were developed to beat Deepfakes in their own way. Generally, the detection problem is considered as a binary classification task, where the two classes are pristine and forged images. 
The forgery detection models usually extract the visual features from the spatial domain and then perform the classification \cite{rossler2019faceforensics++, wang2020video}. In order to further improve the detection performance, some methods exploit various auxiliary features, such as the facial manipulation masks \cite{li2020face}\cite{zhao2021multi}, facial landmarks \cite{li2018exposing} or head directions \cite{yang2019exposing}. 
Besides, the inconsistency in the media source, such as the face-background inconsistency \cite{zhao2021learning} and image-voice inconsistency \cite{mittal2020emotions}, is another cue for forgery detection. 
However, forgery detection solely based on spatial features is found not robust to compression on the source data, which significantly impairs the discrimination performance.
This is because common compression methods such as JPEG or H.264 will remarkably degrade the information in the source media and also contaminate the data in spatial domain by compression errors.
It turns out that forgery artifacts can also be captured in the frequency domain \cite{chen2021local,durall2019unmasking,li2004live,qian2020thinking,wang2020cnn,yu2019attributing}, even though the forgery algorithms is very strong to create the perceptually indistinguishable images and the compression operation further increased the discrimination difficulty.
% Therefore, the detection models that exploit RGB-domain information cannot extract sufficient information or even could be misleaded by the compression noise so that fail to properly discriminate different types of data.
% Problem of SOTA models… too large, low efficiency. So that we need model compression. But experiments shows that existing KD methods do not perform well on the dual-branch distillation.

Forgery detection methods using features from both spatial domain and frequency domain achieved excellent performance\cite{chen2021local, li2021frequency, liu2021spatial, luo2021generalizing}, but at the cost of training and reference efficiency due to their complex multi-branch architectures. 
Knowledge distillation (KD) is a natural choice to improve the efficiency of these models, where the knowledge from a large teacher model guides the learning process of small student networks, in the forms of logits distillation \cite{hinton2015distilling}, intermediate features distillation\cite{romero2014fitnets} and so on.

In this paper, we empirically find that existing knowledge distillation methods tend to lapse when trying to distill information from a two-branch model into a single branch model. Specifically, for a dual-branch network that takes in different types of information and employ spatial-domain and frequency-domain information simultaneously \cite{chen2021local}, the performance of applying knowledge distillation on both branches is even worse than only distilling the information from the RGB branch.
This could attribute to that conflicts exist between the RGB and frequency cues, with diverged directions of loss gradients during training.
Such conflict is not being properly resolved in current distillation methods when merging the two types of knowledge.

% So we propose the two-in-one framework, outperform all the existing methods.
To this end, we propose a novel \textit{two-in-one knowledge distillation} (TOKD) framework for efficient facial forgery detection. The proposed method can not only take advantages of the complementary information from both the spatial and frequency domains but also mediate the contradicting information between them.
In order to efficiently retrieve the complementary information, our framework distills and merges the knowledge from two branches with different inputs into a single-branch student network that only needs the original data as input. 
As for the conflicting part, a rotation module with gradient homogenization is integrated into our framework to reduce the conflicts between the gradients when learning from the RGB and frequency information.

Experimental results on FaceForensics++ \cite{rossler2019faceforensics++} and Celeb-DF \cite{li2020celeb} datasets demonstrate that the proposed framework achieves outstanding performance on the forgery detection task. Specifically, the single-branch student model trained by our TOKD framework achieved the state-of-the-art accuracy on Celeb-DF dataset, and have the best performance among the models with similar size on the FaceForensics++ dataset. Furthermore, the single-branch student model after knowledge distillation even outperforms the dual-branch teacher model, validating the effectiveness of our method in exploiting the complementary information and resolving the contradicting information from different domains.

\section{Related Work}
\label{sec:relate}

\subsection{Facial Forgery Data and Detection}

The facial forgery benefit a lot from the fast development of computer graphics and deep neural networks. The facial forgery data are usually created by manipulating the real medias or generating from scratch through the deep generative models such as the family of generative adversarial networks (GANs) \cite{choi2018stargan,goodfellow2020generative,brock2018large,karras2017progressive,karras2019style}. There are also benchmark datasets consisting of the pristine and synthetic images generated by different kinds of generative models. For example, the \textit{FaceForensics++} dataset \cite{rossler2019faceforensics++} which is a facial forgery dataset contains 1000 video sequences manipulated by the Deepfake\cite{GitHubDFfaceswap}, FaceSwap\cite{GitHubMKfaceswap}, Face2Face\cite{thies2016face2face} and NeuralTextures\cite{thies2019deferred}.  

Many DNN-based forgery detection models have been proposed and achieved promising performance on the forgery discrimination tasks. Most of the methods exploit the information from the spatial-domain such as HSV and RGB. For example, the \cite{huang2020identification}\cite{mccloskey2018detect} try to used color space features for classification. Besides, \cite{li2020face} detect the suspicious artifacts in the face swapping boundary, \cite{zhao2021learning} try to find out the inconsistency between face and background. However, when the forgery data generated by these learning-based models further being compressed by media compression algorithms like H.264, the discrimination performance of these spatial-domain based methods degraded since the compression operation brings loss of spatial-domain information. To this end, models that explore frequency-domain information were explored.

Frequency domain information has been an important role in image classification for a long time \cite{franzen2018image}\cite{sarlashkar1998feature}. In the forgery detection area, many attempts have been made on exploiting frequency cues. To convert the data into the frequency domain and then mine the underlying frequency information, Discrete Fourier Transform (DFT), Discrete Cosine Transform (DCT) and Wavelet Transform are widely used. For example, \cite{durall2019unmasking} makes use of frequency information by applying DFT to transform spatial-domain data into frequency-domain and then averaging the amplitudes of different frequency bands. \cite{qian2020thinking} take advantage of frequency-aware decomposed image components, where the DCT is applied, and local frequency statistics with the help of a two-stream collaborative learning framework to mine the forgery patterns. \cite{chen2021local} also applied DCT for frequency transformation and then use a two-branches architecture where one branch takes the original image as input and the other branch takes the frequency-aware processed image as input for detection. The frequency-aware process includes DCT transformation, high-pass filtering and inverse DCT transformation, which implies a multi-view learning problem \cite{you2017learning, li2017discriminative}. The \cite{chen2021local} claimed to achieve the state-of-the-art performance on widely-used facial detection benchmarks.

\subsection{Knowledge Distillation}
The recent advances in forgery detection widely apply two-branch models for both spatial domain and frequency domain information learning, these models are high in detection accuracy but low in reference efficiency. While there are a large number of forgery media needs to be detected and the edge devices are usually not powerful enough to run large models rapidly, it is important to compress the model and improve the detection efficiency without losing too much performance. Thus, the KD techniques are introduced. The concept of KD was first proposed in \cite{hinton2015distilling}, which is a learning pattern that a larger teacher network use its logits output to guide the logits output of a smaller student network. The mediate feature distillation, as a weakly supervised signal \cite{xu2019positive,xu2014large}, has been demonstrated to work better than the logits \cite{romero2014fitnets,zagoruyko2016paying,ahn2019variational,tung2019similarity}. For example, FitNet \cite{romero2014fitnets} uses a deeper but thinner student network to learn the intermediate representations and output from the teacher network. The student model turns out to outperform the larger teacher with around 10 times fewer parameters. Besides the vanilla feature transfer, feature attention also could be exploited. \cite{zagoruyko2016paying} defined a type of attention for CNN and then let the CNN-based students to mimic the attention maps of the larger teacher network.

However, through our experiments, these KD methods collapse when trying to distill the knowledge from a two-branches model that leverage both spatial domain and frequency domain information. More specifically, for the two-branches network introduced in \cite{chen2021local}, its RGB branch and frequency branch takes the original image and frequency-transformed image as input, respectively. We tried to distill the information from two branches and merge them into a single-branch student. However, this merging distillation has worse performance compared with only distilling the information from the RGB branch. We address this problem by carefully designed feature extractors and the gradient homogenization technique.

\section{Dual-branch Teacher Network}

% \begin{figure*}[t]
%   \centering
%   \includegraphics[width=0.9\linewidth]{latex/img/teacher.png}

%   \caption{Dual-branch Teacher Network}
%   \label{fig:teacher}
% \end{figure*}

\begin{figure}[t]
  \centering
   \includegraphics[width=0.9\linewidth]{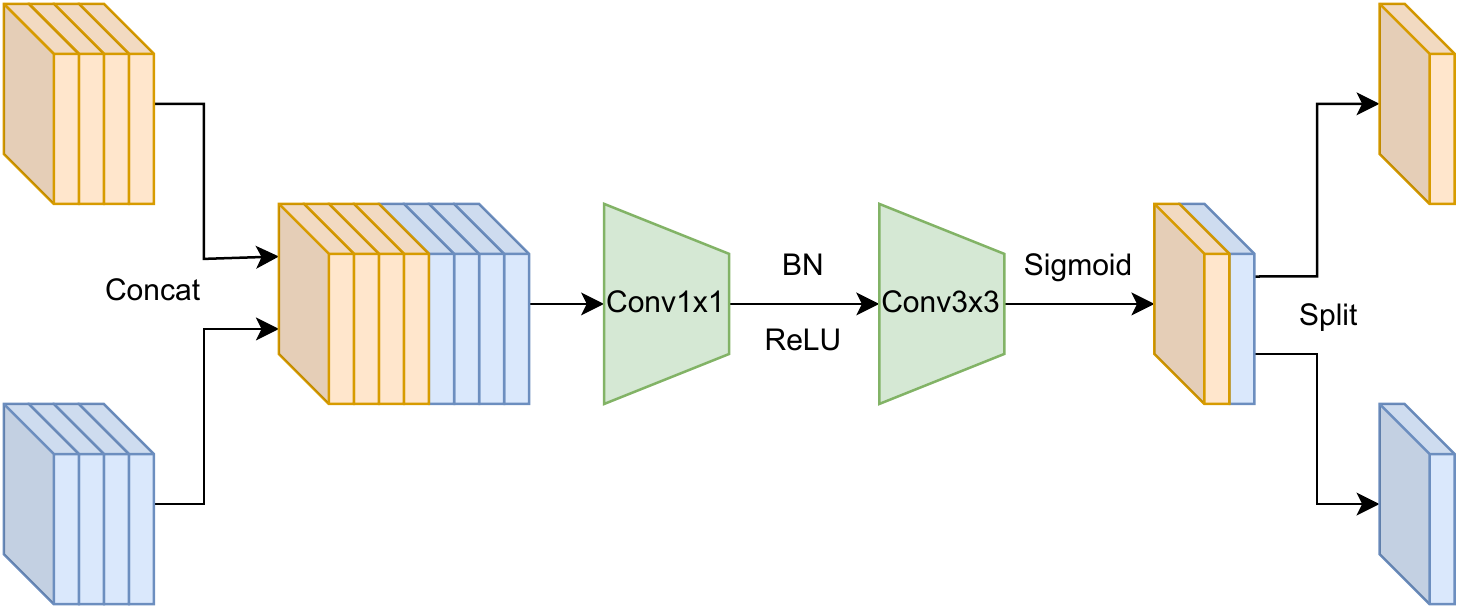}
   \caption{RGB-Frequency Attention Module from \cite{chen2021local}.}
   \label{fig:rfam}
\end{figure}
For the teacher network, we apply a simple dual-branch network architecture, which consists of two branches for learning RGB and frequency information, and a fully connected module in the end to fuse the information. Pre-trained XceptionNets\cite{chollet2017xception} are used as the backbones of each branch. The two branches take in different types of data. Specifically, one branch takes the original images as input and the other branch takes the frequency-aware transformed image as input. These two types of inputs enable the two branches to learn information from the spatial domain and frequency domain, respectively.

For the frequency-aware transformation, the images will first be transformed from the spatial domain to the frequency domain by \textit{Discrete Cosine Transform} (DCT). The transformed frequency representation has a nice layout that all the low-frequency information is located in the top-left corner while all the high-frequency information is located in the bottom-right. According to this, a high-pass filter which will set the top-left triangle area of the feature representation to zero is applied. However, the frequency representation is not feasible for convolution neural networks to process since they do not match the shift-invariance and local consistency owned by natural RGB images. So, in the last step, we apply the \textit{Inverse Discrete Cosine Transform} (IDCT) on the filtered data to get the desired spatial-domain representation that only keeps high-frequency information. Formally, given an input image $\mathbf{x}$, we have:
\begin{equation}
    \mathbf{x}^F = D^{-1}(\mathcal{F}(D(\mathbf{x}))),
\end{equation}
where $D$ and $D^{-1}$ denotes the DCT and IDCT, $\mathcal{F}$ denotes the high-pass filter, $\mathbf{x}^F$ denotes the final data.

In order to fuse the RGB and frequency information to obtain a more comprehensive feature representation, three RGB-Frequency Attention Module (RFAM) ~\cite{chen2021local} are added after the intermediate layers of the backbone. As shown in \cref{fig:rfam}, the RFAM first concatenate the paired internal feature maps from two branches, then pass it through two convolutional modules and then split the final output from the middle by channel to obtain two attention maps for RGB and frequency branch, respectively. These attention maps are used to enhance the RGB and frequency features in the network by multiplication operation.

At the end of the teacher network, the two feature maps from the two branches are concatenated and passed through the pooling and fully connected layers to get the prediction result.

This teacher network can leverage both spatial-domain and frequency-domain information to gain better performance, but it is around three times the model size as the vanilla XceptionNet \cite{chollet2017xception}. This means it is too cumbersome for efficient forgery detection. To improve the model efficiency, knowledge distillation techniques can be applied.
However, in order to make the single-branch model learn from the dual-branch model, the knowledge from the two branches needs to be not only properly distilled, but also smoothly fused.
Through our experimental analysis, previous KD methods \cite{zagoruyko2016paying, passalis2018learning, ahn2019variational, tung2019similarity, romero2014fitnets} perform well on feature distillation but fail on fusing dual-domain knowledge. 
To address this problem, we propose the novel \textit{Two-in-One Knowledge Distillation} framework, which considers both the knowledge extraction and fusion in the dual-branch model distillation tasks.

\section{Method}
\label{sec:method}

% First introduce each components of the framework, then give an overall workflow, then talk about the optimization

\begin{figure}[t]
  \centering
%   \fbox{\rule{0pt}{2in} \rule{0.9\linewidth}{0pt}}
   \includegraphics[width=1.05\linewidth]{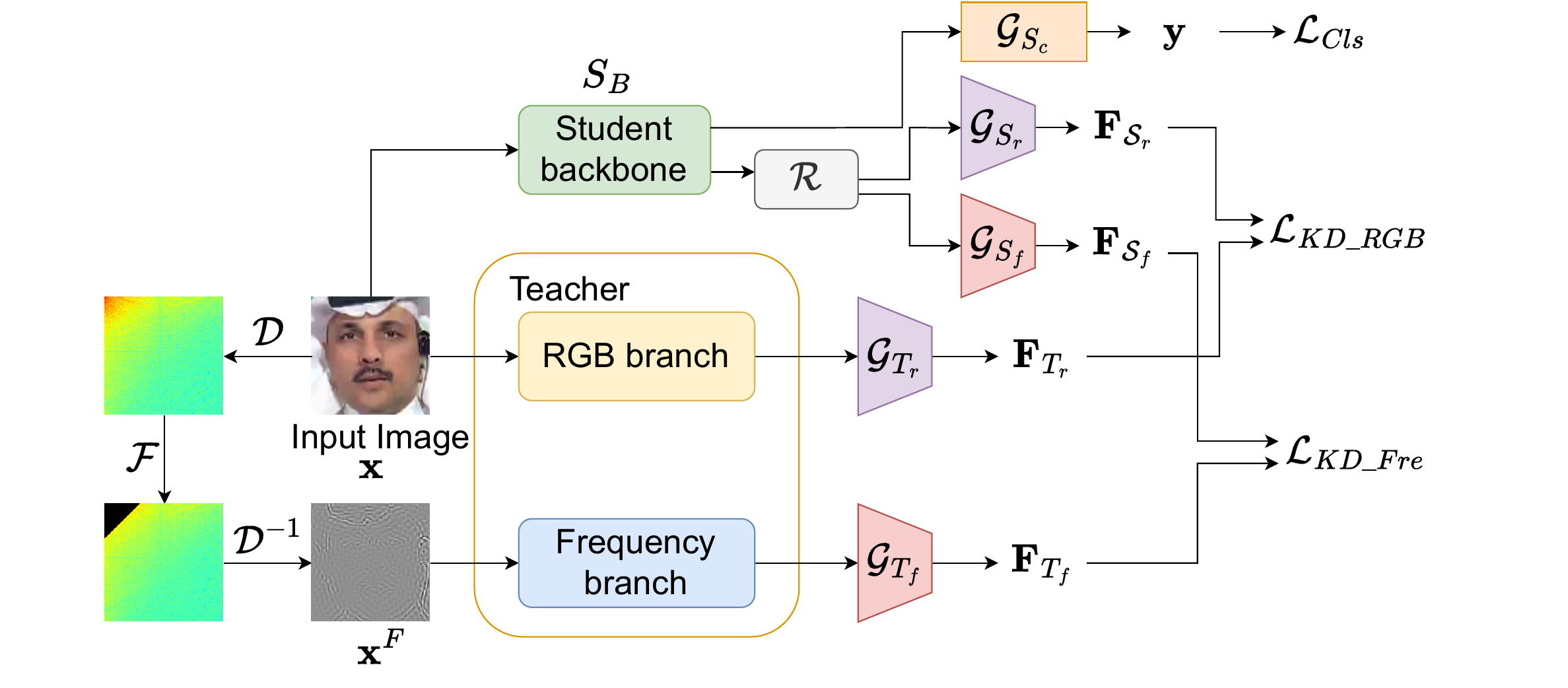}

   \caption{Two-in-One Knowledge Distillation Framework. The student backbone $S_B$ learns the knowledge from the RGB and frequency branches from teacher with the help of dedicated feature projectors and the rotation module $\mathcal{R}$.}
   \label{fig:framework}
\end{figure}

Existing knowledge distillation methods could not be optimal to tackle the fusion of two types of information, as merging the information from two distinct branches (RGB v.s. frequency) directly would bring in conflicts. To solve this problem, we propose the \textit{Two-in-One knowledge distillation} framework which enables a single-branch student network to distill knowledge from a dual-branch teacher network with the help of dedicated feature projectors and the gradient homogenization techniques. The overall pipeline line of  is shown in \cref{fig:framework}.

In this section, we first introduce the feature projectors. Then we introduce the gradient homogenization method we use for resolving the feature conflicts. After this, we formalize the overall training and referencing process of our framework. In the end, we talk about the optimization methods for our framework.

\subsection{Two-in-One knowledge Distillation}

Let $\mathcal{S}$ denote the student network. Besides ground-truth labels,  we expect the student network to simultaneously learn both of the intermediate RGB and frequency feature information from teacher network $\mathcal{T}$. To this end, we first introduce two feature projectors and one classification module after the vanilla convolutional neural network backbone $S_B$. The frequency feature projector $\mathcal{G}_{S_f}$ and RGB feature projector $\mathcal{G}_{S_r}$ are designed to project the feature maps of backbone into spatial domain and frequency domain, respectively, to get the RGB and frequency feature maps, so that they can be guided by the corresponding feature maps from the teacher. $\mathcal{G}_{S_r}$ consists of two convolutions followed by the batch normalization (BN) \cite{ioffe2015batch} and ReLU. $\mathcal{G}_{S_f}$ consists of three convolutions followed by the batch normalization and ReLU as well. 

Formally, let $\mathbf{F}_{S_r}$ denote the extracted RGB feature map and $\mathbf{F}_{S_f}$ denote the extracted frequency feature map from the student network with the input $x$, i.e.,
\begin{equation}
    \mathbf{F}_{S_r} = \mathcal{G}_{S_r}(S_B(\mathbf{x})),
\label{eq:stu-rgb-feat}
\end{equation}
and
\begin{equation}
    \mathbf{F}_{S_f} = \mathcal{G}_{S_f}(S_B(\mathbf{x})).
\label{eq:stu-fre-feat}
\end{equation}
In the teacher network, there are also the corresponding RGB feature projector $\mathcal{G}_{T_r}$ and frequency feature projector $\mathcal{G}_{T_f}$ after the RGB branch and frequency branch. They are designed to extract the condensed feature representation to guide the corresponding feature learning in student network. They have similar architectures as those of $\mathcal{G}_{S_r}$ and $\mathcal{G}_{S_f}$. In particular, $\mathcal{G}_{T_r}$ contains three convolutions followed by the BN and ReLU, while $\mathcal{G}_{T_f}$ contains five convolutions with the BN and ReLU added after the second and fourth layer. Let $\mathbf{F}_{T_r}$ denote the condensed RGB feature map and $\mathbf{F}_{T_f}$ denote the condensed frequency feature map. We have

\begin{equation}
    \mathbf{F}_{T_r} = \mathcal{G}_{T_r}(T_r(\mathbf{x})),
\end{equation}
and
\begin{equation}
    \mathbf{F}_{T_f} = \mathcal{G}_{T_f}(T_f(\mathbf{x}^F)),
\end{equation}
where $T_r$ and $T_f$ represent the RGB branch and frequency branch of the teacher network, respectively.

After getting the desired features from both student and teacher network, the student network learns the RGB information from teacher network by minimizing the distance between $\mathbf{F}_{S_r}$ and $\mathbf{F}_{T_r}$. Formally, we have:

\begin{equation}
  \mathcal{L}_{KD\_RGB} = \mathcal{D}_{RGB}(\frac{\mathbf{F}_{T_{r}}}{||\mathbf{F}_{T_{r}}||_2}, \frac{\mathbf{F}_{S_r}}{||\mathbf{F}_{S_r}||_2}),
  \label{eq:rgb_distill_loss}
\end{equation}
where the $\mathcal{D}_{RGB}$ represents the distance metrics for RGB features. Similarly, the student network learns the frequency information from the teacher by minimizing the distance between $\mathbf{F}_{S_f}$ and $\mathbf{F}_{T_f}$. So, we have:
\begin{equation}
  \mathcal{L}_{KD\_Fre} = \mathcal{D}_{Fre}(\frac{\mathbf{F}_{S_f}}{||\mathbf{F}_{S_f}||_2}, \frac{\mathbf{F}_{T_f}}{||\mathbf{F}_{T_f}||_2})
  \label{eq:fre_distill_loss}
\end{equation}
where the $\mathcal{D}_{Fre}$ is the distance metrics for frequency features.

A classification module $\mathcal{G}_{S_{c}}$ is further used to output the predictions, which will be supervised by the ground-truth label. It contains a pooling layer followed by a fully connected layer, which is a common architecture that classification models usually use in the output layer. 
%It should be noted that the $S_B(\mathbf{x})$ as input directly, rather than rotate it first. 
Formally, we have:
\begin{equation}
    \mathbf{y} = \mathcal{G}_{S_c}(S_B(\mathbf{x})).
    \label{eq:pred}
\end{equation}
The student network learns from the ground-truth label by minimizing the cross-entropy loss between the student network prediction $\mathbf{y}$ and ground-truth label $\mathbf{Y}$. Denoting $CE$ as the cross-entropy loss, we have:
\begin{equation}
    \mathcal{L}_{Cls} = CE(\mathbf{y}, \mathbf{Y})
\label{eq:cls}
\end{equation}

The classification loss in \cref{eq:cls} and the feature distillation losses in \cref{eq:rgb_distill_loss} and \cref{eq:fre_distill_loss} together lead to the final objective function,
\begin{equation}
    \mathcal{L}_{\mathcal{S}} = \mathcal{L}_{Cls} + \alpha_1 \cdot \mathcal{L}_{KD\_RGB} + \alpha_2 \cdot \mathcal{L}_{KD\_Fre},
    \label{eq:net_loss}
\end{equation}
where $\alpha_1$ and $\alpha_2$ are the weight hyper-parameters that need to be tuned.

\subsection{Gradient Homogenization}
\label{subsec:rotation}

As RGB and frequency feature maps are information in spatial domain and frequency domain, respectively, straightforwardly forcing the student network's feature map to be close to RGB and frequency feature maps from the teacher network at the same time would increase the burden of knowledge distillation and could result in an inferior student network that neither properly learns the RGB information nor the frequency information. 

According to \cref{eq:rgb_distill_loss} and \cref{eq:fre_distill_loss}, the direction of the gradients coming from these distillation losses to update $S_{B}$ can be diverged. To mitigate the potential gradient conflict from these two distillation losses, we introduce a rotation module to perform domain-specific rotations on the output of $S_B$, i.e., 
\begin{equation}
    \mathbf{F}_{S_r} = \mathcal{G}_{S_r}(R_r(S_B(\mathbf{x}))),
\label{eq:stu-rgb-feat}
\end{equation}
and
\begin{equation}
    \mathbf{F}_{S_f} = \mathcal{G}_{S_f}(R_f(S_B(\mathbf{x}))),
\label{eq:stu-fre-feat}
\end{equation}
where the $R_r$ and $R_f$ are the rotation matrices in the rotation module. The rotation module $\mathcal{R}$ contains two optimizable rotation matrices $R_r, R_f \in SO(d)$, where $SO(d)$ denote the the special orthogonal group with matrices of dimension $d$. The $d$ is a parameter that need to be tuned. Before the output of student backbone being passed to $\mathcal{G}_{S_r}$ and $\mathcal{G}_{S_f}$, it will be first passed through the rotation module $\mathcal{R}$. $R$ rotates the $S_B(\mathbf{x})$ by multiplying it with $R_r$ and $R_f$ to get the rotated RGB feature $R_r(S_B(\mathbf{x}))$ and frequency feature $R_f(S_B(\mathbf{x}))$, respectively.

% As we mentioned before, merging information from both RGB and frequency branch into a single branch brings optimization conflicts. Thus, we introduce the rotation module which will rotate the last feature of student backbone to reduce the conflict between the gradients of RGB and frequency information learning.

Noted that the learnable rotation matrices are optimized to eliminate the direction conflict between the gradients of RGB loss and frequency loss, rather than to reduce the distillation loss directly. Let $k \in \{r, f\}$, that is, $k$ can stand for either RGB (r) or frequency (f). The way we optimize the rotation matrices is by minimizing:
\begin{equation}
    \mathcal{L}_{R_k} = -\sum_n\langle R_k^{\intercal}\nabla_{\mathbf{m}_k} \mathcal{L}(\mathbf{F}_{S_k}, \mathbf{F}_{T_k}), g_n\rangle,
    \label{eq:rotation_optim}
\end{equation}
where $\langle,\rangle$ denotes the Cosine Similarity. $\mathbf{m}_k$ is the rotated feature map, that is:
\begin{equation}
    \mathbf{m}_k = R_k(S_{B}(\mathbf{x})).
\end{equation}
And $g_n$ is the target vector which points to the direction that we want the gradients from both RGB and frequency distillation to follow. We define:
\begin{equation}
    g_n = \frac{1}{2}(v^{r}_n + v^{f}_n),
\end{equation}
where $v^{r}_n$ and $v^{f}_n$ denotes the RGB gradient and frequency gradient of one sample in the batch. This means $g_n$ is the average gradient of RGB and frequency gradients.

As a result, in the training process of our TOKD framework, we need to optimize \cref{eq:net_loss} and \cref{eq:rotation_optim} alternatively. Specifically, in each iteration, the gradient from loss $\mathcal{L}_\mathcal{S}$ will be used to update the parameters of the student backbone $S_B$ and all the projectors in the same manner as normal KD methods. Then the gradient  $\nabla_{R_k}\mathcal{L}_{R_k}$ calculated from loss $\mathcal{L}_{R_k}$ will update the rotation matrix $R_k$ by:
\begin{equation}
    R_k = R_k - \eta_{\mathcal{R}}\nabla_{R_k}\mathcal{L}_{R_k},
\end{equation}
where $\eta_\mathcal{R}$ denotes the learning rate for rotation module.

\subsection{Training and Prediction Process}
During the training process, given an image $\mathbf{x}$, it will first be passed through the backbone of the student network to get the backbone feature $S_B(\mathbf{x})$. Then the rotation module will rotate $S_B(\mathbf{x})$ by multiplying it with two rotation matrices and two rotated feature maps $R_r(S_B(\mathbf{x}))$ and $R_f(S_B(\mathbf{x}))$ will be obtained. By passing the rotated feature maps through the corresponding feature projector. We will get the RGB feature map $\mathbf{F}_{S_r}$ and frequency feature map $\mathbf{F}_{S_f}$. 

For the teacher network, the input data $\mathbf{x}$ and $\mathbf{x}^F$ will be passed through RGB and frequency branch separately, and then the $\mathcal{G}_{T_r}$ and $\mathcal{G}_{T_f}$ will be applied to get the condensed feature representations $\mathbf{F}_{T_r}$ and $\mathbf{F}_{T_f}$. These two feature maps will guide the $\mathbf{F}_{S_r}$ and $\mathbf{F}_{S_f}$ from student network according to \cref{eq:rgb_distill_loss} and \cref{eq:fre_distill_loss}.The $S_B(\mathbf{x})$ will be fed into classification module directly for the classification output which will be guided by the ground-truth label according to \cref{eq:cls}.

% Modify this 
This optimization problem of our framework can be considered as a Stackelberg game, which is a two-player game that the leader and follower move alternatively trying to minimize their own losses. In our case, the leader is the optimizer of $\mathcal{R}$ and the follower is the optimizer for the $S_B$ together with its following projectors. The leader knows how the follower will response to their move. We can ensure the convergence of both loss \cref{eq:net_loss} and \cref{eq:rotation_optim} by following the \cite{fiez2020implicit}, which is to make the optimizer of rotation matrix (leader) optimize slower than the optimizer of student network (follower). In other words, let $\eta_\mathcal{S}$ denotes the learning rate for student backbone $S_B$ and all the projectors after it, both the $\mathcal{L}_{\mathcal{S}}$ and $\mathcal{L}_{\mathcal{R}}$ will be optimized to a local optimum as long as the $\eta_\mathcal{R}$ decrease faster than the $\eta_\mathcal{S}$ \cite{javaloy2021rotograd}.

During the prediction, the rotation module $\mathcal{R}$ and the two feature projectors $\mathcal{G}_{S_r}$ and $\mathcal{G}_{S_f}$ will be removed. Only the student backbone $\mathcal{S}_B$ and the classification module $\mathcal{G}_{S_c}$ will be used. The input image $\mathbf{x}$ will first be passed through the $S_B$ and then be passed through the $\mathcal{G}_{S_c}$ to get the prediction output $\mathbf{y}$, as formally defined in \cref{eq:pred}.

\begin{table}
  \centering
  \begin{adjustbox}{width=\linewidth}
    \begin{tabular}{l c c c c}
    \toprule
    & ResNet18-T & ResNet34-T & Xception & Teacher\\
    \midrule
     \#Params & 2.8 M & 5.3 M & 22.8 M & 62.8 M \\
     GFLOPs & 0.91 & 1.78 & 8.45 & 20.14 \\
    \bottomrule
  \end{tabular}
  \end{adjustbox}
  \caption{Number of parameters and GFLOPs (Giga floating point operations) for each networks we use in our experiments.}
  \label{tab:num-parameters}
\end{table}

\begin{table}
  \centering
  % \begin{tabular}{@{}lc@{}cc@{}cc@{}}
  \begin{adjustbox}{width=1\linewidth}
    \begin{tabular}{l l c c c c c c}
    \toprule
    & & \multicolumn{2}{c}{FF++(c40)} & \multicolumn{2}{c}{Celeb-DF}\\
    Method & \#Param & Acc(\%) & AUC(\%) & Acc(\%) & AUC(\%) \\
    \midrule
    Multi-task \cite{nguyen2019multi} & - & 81.30 & 75.59 & - & - \\
    Xception \cite{chollet2017xception} & 22.8 M & 86.86 & 89.30 & 97.90 & 99.73 \\
    Two-branch \cite{masi2020two} & - &  86.34 & 86.59 & - & - \\
    SPSL \cite{liu2021spatial} & - & 81.57 & 82.82 & - & - \\
    RFM \cite{wang2021representative} & 22.8 M & 87.06 & 89.83 & 97.96 & 99.94 \\
    % Freq-SCL \cite{li2021frequency} & & & 89.00 & 92.39 & - & - \\
    AddNet \cite{zi2020wilddeepfake} & -  & 87.50 & 91.01 & 96.93 & 99.55 \\
    Face X-ray \cite{li2020face} & 26.2 M & - & 61.60 & - & - \\
    $F^3$-Net \cite{qian2020thinking} & 48.3M & \textbf{90.43} & 93.30 & 95.95 & 98.93 \\
    MultiAtt \cite{zhao2021multi} & 417.6 M & 88.69 & 90.40 & 97.92 & 99.94 \\
    % RECCE & & & 91.03 & 95.02 & 98.59 & 99.94\\
    TOKD(Ours) & \textbf{22.8 M} & 87.78 & 89.59 & \textbf{98.73} & 98.45 \\
    \bottomrule
  \end{tabular}
  \end{adjustbox}
  \caption{The pre-trained Xception trained by our TOKD framework performs favorably over most of current state-of-the-art approaches, while use much less parameters.}
  \label{tab:compare-detection}
  \vskip -0.2in
\end{table}

\section{Experiment}

\begin{table*}
  \centering
  % \begin{tabular}{@{}lc@{}cc@{}cc@{}}
  \begin{adjustbox}{width=1\textwidth}
    \begin{tabular}{l c c c c c c c c c c c c c c c c c c c c c c }
    \toprule
    & & & \multicolumn{2}{c}{TOKD(Ours)} & \multicolumn{2}{c}{AT \cite{zagoruyko2016paying}} & \multicolumn{2}{c}{PKT \cite{passalis2018learning}} & \multicolumn{2}{c}{VID \cite{ahn2019variational}} & \multicolumn{2}{c}{SP \cite{tung2019similarity}} & \multicolumn{2}{c}{FitNet \cite{romero2014fitnets}}\\
    Dataset & Student & Vanilla & RGB & Both  & RGB & Both & RGB & Both & RGB & Both  & RGB & Both & RGB & Both\\
    \midrule
    & Res18-T & 80.96 & 81.67 & 82.45$\textcolor{green}{\uparrow}$ & 81.45 & 81.26$\textcolor{red}{\downarrow}$ & 81.42 & 81.29$\textcolor{red}{\downarrow}$ & 83.73 & 83.10$\textcolor{red}{\downarrow}$ & 81.41 & 81.16$\textcolor{red}{\downarrow}$ & 82.46 & 82.23$\textcolor{red}{\downarrow}$ \\
    FF++& Res34-T & 81.10 & 82.32 & 82.94$\textcolor{green}{\uparrow}$ & 81.68 & 81.58$\textcolor{red}{\downarrow}$ & 81.60 & 81.51$\textcolor{red}{\downarrow}$ & 83.52 & 82.51$\textcolor{red}{\downarrow}$ & 81.70 & 81.21$\textcolor{red}{\downarrow}$ & 83.20 & 81.95$\textcolor{red}{\downarrow}$ \\
    & Xception & 82.47 & 83.76 & 85.16$\textcolor{green}{\uparrow}$ & 82.60 & 81.90$\textcolor{red}{\downarrow}$ & 82.71 & 82.46$\textcolor{red}{\downarrow}$ & 85.04 & 84.67$\textcolor{red}{\downarrow}$ & 82.95 & 82.44$\textcolor{red}{\downarrow}$ & 85.55 & 84.99$\textcolor{red}{\downarrow}$ \\
    \midrule
    & Res18-T & 93.22 & 95.82 & 97.23$\textcolor{green}{\uparrow}$ & 95.84 & 95.68$\textcolor{red}{\downarrow}$ & 93.18 & 93.18$\textcolor{red}{\downarrow}$ & 96.51 & 95.84$\textcolor{red}{\downarrow}$ & 94.85 & 94.61$\textcolor{red}{\downarrow}$ & 96.45 & 95.28$\textcolor{red}{\downarrow}$  \\
    CelebDF & Res34-T & 93.61 & 96.12 & 96.56$\textcolor{green}{\uparrow}$ & 96.49 & 96.12$\textcolor{red}{\downarrow}$ & 94.63 & 94.27$\textcolor{red}{\downarrow}$ & 96.97 & 96.28$\textcolor{red}{\downarrow}$ & 95.44 & 94.67$\textcolor{red}{\downarrow}$ & 96.20 & 96.12$\textcolor{red}{\downarrow}$ \\
    & Xception & 94.17 & 97.74 & 98.22$\textcolor{green}{\uparrow}$ & 96.97 & 96.21$\textcolor{red}{\downarrow}$ & 97.38 & 97.15$\textcolor{red}{\downarrow}$ & 96.91 & 95.90$\textcolor{red}{\downarrow}$ & 97.92 & 97.88$\textcolor{red}{\downarrow}$ & 96.37 & 95.92$\textcolor{red}{\downarrow}$ \\
    \bottomrule
  \end{tabular}
  \end{adjustbox}
  \caption{The performance of different knowledge distillation methods on FF++(c40) and Celeb-DF dataset. The non-pretrained XceptionNet trained by our TOKD framework performs favorably over most of current state-of-the-art approaches, while use much less parameters.}
  \label{tab:conflict}
\end{table*}

\subsection{Datasets}
Two facial forgery datasets are used for this section, which are the FaceForensics++ (FF++)\cite{rossler2019faceforensics++} and the Celeb-DF \cite{li2020celeb}. FF++ collects 1,000 real videos from YouTube and generates facial forgery data with four generation methods which are Deepfake \cite{GitHubDFfaceswap}, FaceSwap \cite{GitHubMKfaceswap}, Face2Face \cite{thies2016face2face} and NeuralTextures \cite{thies2019deferred} respectively. It generates 1,000 forgery videos for each generation method; thus it includes 5,000 videos in total (1000 real videos + 4,000 forgery videos). On the other hand, Celeb-DF provides a large forgery and more challenging dataset that contain 590 YouTube videos covering different age and gender groups and generate 5639 face swapping videos.

As the preprocessing step, we apply the H.264 video compression method for the FF++ dataset, as the model trained with raw FF++ data can approach 99\% accuracy easily. Specifically, each FF++ video is compressed with C40 (1:40) compression ratio. In addition, we extract 32 frames from each video for both FF++ and Celeb-DF videos and resize them into 299x299 pixels.

\subsection{Experiment settings}
We selected XceptionNet \cite{chollet2017xception} (pre-trained on ImageNet) as the backbone network for each branches of the teacher network. In in our experiment settings, we use Adam \cite{kingma2014adam} optimizer with default parameters ($\beta_1 = 0.9$, $\beta_2 = 0.999$). The learning rate is started with 0.0001 for training on FF++, and 0.001 for training on celeb-DF. A step learning rate scheduler is applied with step=5 and $\gamma=0.1$. The total training epoch is 15, and the batch size is 64. For the weight parameters $\alpha_1$ and $\alpha_2$ in \cref{eq:net_loss}, the different value combinations do not have a huge difference. The optimal values are $\alpha_1=\alpha_2=10$ for FF++ and $\alpha_1=\alpha_2=200$ for celeb-DF. The optimal dimension value $d$ of the rotation matrices is 1024 on both datasets. We evaluate the model performance at the frame level, using Accuracy (Acc) and Area Under Curve (AUC) metrics. During the training progress, We select the best models based on the validation accuracy and we report the performance results in the testing set in the following parts. In addition, we applied default parameter settings for other SOTA models and algorithms. What's more, we selected three different versions of student models for the experiment, which are the ResNet18-thin, ResNet34-thin and XceptionNet, respectively. The ResNet18-thin and ResNet34-thin are the thinner version of ResNet18 and ResNet34 \cite{targ2016resnet} by cutting their channel numbers into half in each layer. The detailed number of parameters and floating point operations (FLOPs) needed for processing a single input for these models are shown in the following \cref{tab:num-parameters}.

\begin{table}
  \centering
  % \begin{tabular}{@{}lc@{}cc@{}cc@{}}
    \begin{tabular}{l c c c c}
    \toprule
    & \multicolumn{2}{c}{FF++(c40)} & \multicolumn{2}{c}{Celeb-DF} \\
    Model & Acc(\%) & AUC(\%) & Acc(\%) & AUC(\%) \\
    \midrule
    Xception & 86.86 & 89.30 & 97.90 & 99.73 \\
    Teacher & 87.53 & 89.47 & 98.65 & 98.13 \\
    Student &  87.78 & 89.59  & 98.73 & 98.45 \\
    \bottomrule
  \end{tabular}
  \caption{Student network can outperform teacher network after training by our TOKD framework. Both teacher and student network use pre-trained XceptionNet as backbone.}
  \label{tab:compare-teacher}
  \vskip -0.2in
\end{table}

\subsection{Experiment Results}
The \cref{tab:conflict} compares the performance of our model with other knowledge distillation algorithms. The results indicate that using previous knowledge distillation methods only to transfer RGB-related knowledge from the RGB branch of the teacher to the student can improve students' performance. However, if we use the same algorithm to transfer frequency-related knowledge from the teacher's frequency branch to the student (together with the RGB knowledge transformation), it causes significant down gradation within all previous knowledge distillation methods. Specifically, the student detection accuracy will decrease by 0.43\% for FF++ and 0.52\% for celeb-DF after learning knowledge from both branches. It highlights the potential conflicts of learning RGB and frequency information simultaneously. Thus we applied the gradient homogenization method to align the gradients. Our proposed method significantly increases the performance from RGB-only distillation by 0.93\% accuracy improvement for FF++ and 0.78\% increment for celeb-DF.

Besides, our proposed method gains high performance while keeping a small size. The \cref{tab:compare-detection} presents the size and detection performance compared with our model and previous SOTA facial forgery detection models. The pre-trained XceptionNet is used as our student model. According to the result, our student model has approached the SOTA accuracy on celeb-DF, which is a 0.85\% accuracy improvement on average, which is even higher than the models with complicated architecture, such as F3-Net \cite{qian2020thinking} (2.78\% accuracy improvement with 47\% of size) and MultiAtt \cite{zhao2021multi} (0.81\% accuracy improvement with 0.05\% of size.) Furthermore, we also approach the SOTA performance for FF++(c40) within the models with similar sizes. It has a 0.92\% accuracy increment ahead of the training of XceptionNet itself and a 0.72\% accuracy improvement compared with RFM \cite{wang2021representative}. Compared with the $F^3$-Net which achieves 90.43\% accuracy on FF++(c40), our model only uses half the parameters, which is more efficient, while also having promising performance.

Meanwhile, our proposed method's single brunch student model can outperform the two brunch teacher model. As the \cref{tab:compare-teacher} shows, the student with a pre-trained XceptionNet model achieves 0.25\% and 0.08\% accuracy increment for detecting FF++ and celeb-DF, compared with the teacher model. The result could indicate the recent complicated multiple brunch models are over-parameterized and highlight the importance of reducing model size with model compression methods like our proposed method.

\begin{table}
  \centering
    \begin{tabular}{l c c c c c}
    \toprule
    Rotation $d$ & 128 & 256 & 512 & 1024 & 2048 \\
    \midrule
    FF++(c40) & 80.28 & 80.36 & 81.90 & 82.94 & 80.44 \\
    \bottomrule
  \end{tabular}
  \caption{Accuracy of student models trained under different rotation matrix size on the FF++(c40) dataset. Student model is ResNet18-Thin. We set $\alpha_1=\alpha_2=10$.}
  \label{tab:rotation_size}
  \vskip -0.2in
\end{table}

% \begin{table}
%   \centering
%     \begin{tabular}{l c}
%     \toprule
%     Rotation $d$ & FF++(c40)\\
%     \midrule
%     128 & 80.28  \\
%     256 & 80.36 \\
%     512 & 81.90 \\
%     1024 & 82.94 \\
%     2048 & 80.44 \\
%     \bottomrule
%   \end{tabular}
%   \caption{Accuracy of student models trained under different rotation matrix size. Student model is ResNet18-Thin. $\beta$ is fix to 10.}
%   \label{tab:rotation_size}
% \end{table}

\subsection{Cross-dataset Testing}
To evaluate the generalization ability of the student model trained by our TOKD framework, the cross-dataset testing experiments are conducted. The pre-trained XceptionNet is used as the student backbone. We first train the student by TOKD method on the FF++(c40) dataset, and then test it on the Celeb-DF data set. The reason of this setting is because the highly compressed FF++ dataset is more challenging compared with the raw Celeb-DF. Thus, the student trained on the harder dataset is expected to have better generalization ability. The results in \cref{tab:cross} shows that the student trained by TOKD have superior performance on generalization. The only method with 1.04\% higher AUC than us, which is the MultiAtt \cite{zhao2021multi}, have around 394.8 M more parameters than our student model.

\subsection{Parameters Tuning}
\begin{figure}[t]
  \centering
%   \fbox{\rule{0pt}{2in} \rule{0.9\linewidth}{0pt}}
   \includegraphics[width=0.9\linewidth]{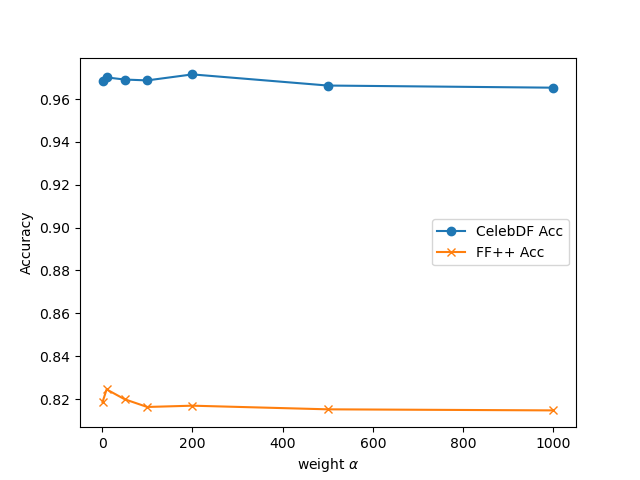}

   \caption{Sensitivity to $\alpha$ of our TOKD method on FF++(c40) and Celeb-DF dataset. ResNet18-thin is used as the student model.}\vspace{-0.2in}
   \label{fig:weight_acc}
\end{figure}

The \cref{tab:rotation_size} demonstrate the influence of the size $d$ of rotation matrices on model performance. We tuned the $d$ from 128 to 2048. The results show that as the $d$ increases in a reasonable range, the student performance will also increase. However, when the $d$ is too large, the performance will drop.

In \cref{eq:cls} we introduced the hyper-parameters $\alpha_1$ and $\alpha_2$, which are the weights for RGB and frequency distillation loss. We find the different value combination of $\alpha_1$ and $\alpha_2$ do not have a noticeable difference in performance, and the optimal values we get are $\alpha_1=\alpha_2=10$ on FF++(c40) and $\alpha_1=\alpha_2=200$ on Celeb-DF. So we set $\alpha_1=\alpha_2=\alpha$, and conduct experiments using different values of $\alpha$ range from 1 to 1000 on both FF++(c40) and Celeb-DF with ResNet18-thin as the student. The results in \cref{fig:weight_acc} indicate the robust performance over such $\alpha$ range. And the $\alpha=10$ on FF++ and the $\alpha=200$ on Celeb-DF have slightly better performance.

\subsection{Ablation Studies and Discussion}

\subsubsection{Effects of distillation modules}
We investigate the impact of different distillation modules in our methods on detection accuracy. According to the \cref{tab:ablation}, it will cause fewer improvements for frequency-only distillation than the RGB-only one (0.46\% vs 1.29\% in FF++). One reason could be the large input domain gap between the frequency input teacher and the RGB input student. Meanwhile, just like we discussed in \cref{tab:conflict}, simply learning two kinds of knowledge by adding their distillation loss cannot improve the performance but is even worse than the frequency-only version, causing 0.23\% down-gradation. Finally, the table highlights the critical roles of projectors and the rotation module to achieve the best performance. 

\subsubsection{Visualization on gradient similarity}
\begin{figure}[t]
  \centering
%   \fbox{\rule{0pt}{2in} \rule{0.9\linewidth}{0pt}}
   \includegraphics[width=0.85\linewidth]{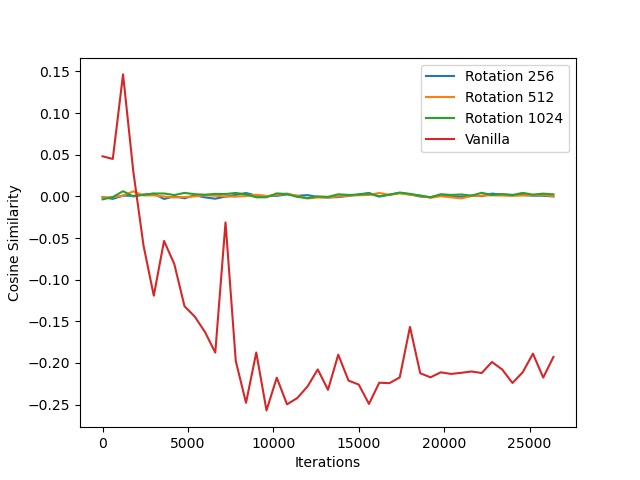}

   \caption{The variation of cosine similarity between RGB and frequency gradients during training under different size $d$ of rotation matrix on FF++(c40) dataset. The gradients have large divergence before applying the rotation module.}   
   \label{fig:gradient_similarity}
\end{figure}

As mentioned before, the gradients of RGB and frequency loss can have direction conflicts. To visualize the similarity between the RGB and the frequency gradients, experiments of applying rotation modules with different sizes $d$ are conducted on the FF++(c40) dataset, where the XceptionNet is used as the student backbone. The \cref{fig:gradient_similarity} presents the similarity between the gradient calculated from $\mathcal{L}_{KD\_RGB}$ and $\mathcal{L}_{KD\_Fre}$. Without applying the gradient rotation method, the two gradients can vastly diverge from each other. The cosine similarity at around -0.2 indicates that the RGB and frequency gradients can point in opposite directions. Such severe direction inconsistency can cause conflicts in the optimization process and can degrade the detection performance of the student model. After applying the rotation method, the level of divergence between two gradients is decreased. The similarity between the rotated gradients is around 0, which means the two gradients after rotation are orthogonal to each other. Although the rotation module does not change the gradients of RGB and frequency distillation losses to the same direction, which is actually hard as they come from different domains,  we can at least ensure that the RGB and frequency gradients do not have a negative effect on each other during the optimization process. And this directly results in the improvement of the detection performance of the student model. 

\label{sec:experiment}

% \begin{figure}[t]
%   \centering
% %   \fbox{\rule{0pt}{2in} \rule{0.9\linewidth}{0pt}}
%   \includegraphics[width=1\linewidth]{latex/img/feature.png}

%   \caption{Visualization of feature maps extracted by Vanilla Xception and the TOKD-trained Xception.}
%   \label{fig:feature}
% \end{figure}

\begin{table}
  \centering
  \begin{adjustbox}{width=1\linewidth}
    \begin{tabular}{c c c c c c}
    \toprule
    RGB & Fre & Projector & Rotation & FF++(c40) & Celeb-DF \\
    \midrule
    & & & & 82.47 & 94.17 \\
    \checkmark & & \checkmark & & 83.76 & 97.74 \\
    & \checkmark & \checkmark & & 82.93& 97.25 \\
    \checkmark & \checkmark & \checkmark & & 82.70 & 97.06  \\
    \checkmark & \checkmark & & \checkmark & 81.21 & 94.84 \\
    \checkmark & \checkmark & \checkmark & \checkmark &  85.16 & 98.22\\
    \bottomrule
  \end{tabular}
 \end{adjustbox}
  \caption{Ablation study on the effectiveness of each model components. Non-pretrained XceptionNet is used as student network.}
  \label{tab:ablation}
\end{table}

\begin{table}
  \centering
    \begin{tabular}{l c c}
    \toprule
    & \multicolumn{2}{c}{Celeb-DF} \\
    Methods & AUC(\%) & EER(\%) \\
    \midrule
    Xception \cite{chollet2017xception} & 61.80 & 41.73 \\
    Add-Net \cite{zi2020wilddeepfake} & 65.63 & 38.54 \\
    MultiAtt \cite{zhao2021multi} & 67.02 & 37.09\\
    RFM \cite{wang2021representative} & 65.63 & 38.54\\
    TOKD(Ours) & 65.98 & 40.34\\
    \bottomrule
  \end{tabular}
  \caption{Cross-dataset testing for evaluation of generalization ability. The student model use XceptionNet as backbone. It is trained on FF++(c40) using TOKD method and tested on Celeb-DF.}
  \label{tab:cross}\vspace{-0.2in}
\end{table}

\section{Conclusion}
\label{sec:conclusion}
In this paper, we propose a novel two-in-one knowledge distillation framework which can smoothly distill and merge both of the spatial and frequency domain knowledge from a dual-branch teacher into a single-branch student network. Extensive experiments shows that the student model trained by our method have superior performance while use much less parameters.

%%%%%%%%% REFERENCES
{\small
\bibliographystyle{ieee_fullname}
\bibliography{main}
}

\end{document}